\documentclass[runningheads]{llncs}

\usepackage{graphicx}
\usepackage{amssymb}
\usepackage{amsmath,tabularx}
\usepackage{xcolor}
\usepackage{multirow}
\usepackage{siunitx}
\usepackage{booktabs}
\usepackage{etoolbox}
\usepackage{color}
\usepackage{lipsum}
\usepackage{tabu}
\usepackage{dsfont}
\usepackage{subcaption,siunitx,booktabs}
\usepackage[labelfont=bf]{caption}

\definecolor{ben}{rgb}{0.9,0.,0.5}

\definecolor{seongtae}{rgb}{0.5,0.,0.5}

\definecolor{todo}{rgb}{1.0, 0., 0.}

\newcommand\asteriskfill{\leavevmode\xleaders\hbox{$\ast\ $}\hfill\kern0pt}
\newcommand{\comment}[1]{}
\begin{document}

\robustify\bfseries
        
\title{OperA: Attention-Regularized Transformers \\ for Surgical Phase Recognition}

\titlerunning{OperA: Attention-Regularized Transformers}

\author{\asteriskfill \\ \asteriskfill}
\author{Tobias Czempiel\inst{1}, Magdalini Paschali\inst{1}, Daniel Ostler\inst{2}, Seong Tae Kim\inst{1}, Benjamin Busam\inst{1}, Nassir Navab\inst{1,3}}

\authorrunning{Czempiel et al.}

\institute{
 Computer Aided Medical Procedures, Technische Universit{\"a}t M{\"u}nchen, Germany
 \and
 MITI, Klinikum Rechts der Isar, Technische Universit{\"a}t M{\"u}nchen, Germany
 \and
 Computer Aided Medical Procedures, Johns Hopkins University, Baltimore, USA 
}

\maketitle              
\begin{abstract}

In this paper we introduce OperA, a transformer-based model that accurately predicts surgical phases from long video sequences. A novel attention regularization loss encourages the model to focus on high-quality frames during training. Moreover, the attention weights are utilized to identify characteristic high attention frames for each surgical phase, which could further be used for surgery summarization. OperA is thoroughly evaluated on two datasets of laparoscopic cholecystectomy videos, outperforming various state-of-the-art temporal refinement approaches.

\end{abstract}

\section{Introduction} 

Surgical workflow analysis is a crucial task for the operating room (OR) of the future~\cite{Maier-Hein2018}. Specifically, automatic detection of surgical phases is one of its most essential components. An efficient surgical phase recognition system will build the foundation for automated surgical assistance and cognitive guidance~\cite{Garrow2020,Padoy2019}. Online analysis during an ongoing intervention can provide feedback to surgeons and alarm the staff in case of erroneous or adverse events~\cite{Huaulme2020}. Additionally, extracting surgical phases during an operation and group different procedures based on their unique characteristics plays an important role for modern surgical training. 
Since automatic extraction of surgical phases is particularly challenging, advanced Machine Learning (ML) methodologies~\cite{Twinanda2017} have been employed towards solving it.
However, factors such as variability of patient anatomy, surgeon style~\cite{Funke2019} as well as limited availability and quality of training data present problems for modern ML algorithms.

A recent development in ML that could help overcome these challenges in surgical workflow analysis is transformer networks~\cite{Vaswani2017}. Transformers have shown their vast potential for sequential modeling in Natural Language Processing (NLP)~\cite{Devlin2019} and have quickly become the gold standard in this area. 
Transformer networks have the capability to create temporal relationships between current and previous frames using self-attention, much like frequently used LSTM methods~\cite{doi:10.1162/neco.1997.9.8.1735}. However, self-attention enables learning in long sequences without forgetting of previous information which often hampers LSTM-based methods.

An additional advantage of transformer networks and self-attention over other approaches used for surgical phase recognition is their ability to visualize the attention weights for a sequence, which could yield further insights into the decision-making process of a model. 
\subsection{Related Work} 

Automatic extraction of surgical phases was initially performed using binary surgical signals~\cite{Ahmadi2006}, where a comparison with an average surgery determined the surgical phase.
Hidden Markov Models (HMM), provided an extension of this idea capable of online predictions~\cite{Padoy2012}.
In EndoNet, Twinanda et al.~\cite{Twinanda2017} utilized image features extracted with a Convolutional Neural Network (CNN) to predict the surgical phase and surgical tool presence directly from surgical images.
EndoLSTM~\cite{Twinanda2017a} additionally performed temporal refinement with LSTMs~\cite{doi:10.1162/neco.1997.9.8.1735}, which improved the results substantially.
A variety of works combined pre-trained CNNs as feature extractors, followed by temporal refinement with LSTMs~\cite{Yengera2018,Jin2018}.
In MTRCNet-CL~\cite{Jin2020}, Jin et al. proposed a CNN/LSTM model to refine the prediction over short sequences in an end-to-end fashion including a correlation loss to identify phase and tool correlations in an explicit manner.
Czempiel et al.~\cite{tecno} proposed TeCNO, which combined Temporal Convolutional Networks (TCN) with a ResNet-50~\cite{He2016} feature extractor. 
Transformer models were first introduced for NLP~\cite{Vaswani2017} where they quickly became the state-of-the-art in a plethora of downstream tasks~\cite{Brown2020,Devlin2019}.
Furthermore, the versatility of transformers has been showcased not only for vision tasks such as image classification~\cite{Dosovitskiy2020} and text-to-image generation~\cite{Ramesh2021} but also in biology for the challenging protein folding problem with \textit{AlphaFold}~\cite{Heo2020}.
In surgical data sciences, transformers have been explored only for surgical tool \cite{kondo2020lapformer} classification.

An additional aspect of transformers and self-attention is the fact that their attention weights could be used for model insights and explanation. Some works have claimed that attention has limited explanation capabilities~\cite{Jain2019}. However, this assumption has been challenged~\cite{wiegreffe2019} suggesting
that each work should define their notion of explanation since it could be dependent on the task at hand.

In this paper, we introduce for the first time, OperA, a transformer-based method for online surgical phase prediction for laparoscopic operations.
Our contributions are:
\begin{itemize}
    \item[$\circ$] We successfully leverage a transformer-based model for surgical phase recognition that outperforms other temporal refinement methods.
    \item[$\circ$] We propose a novel attention regularizer, that improves the automatic extraction of the most relevant frames with high feature-quality.
    %focus towards features of high quality.
    \item[$\circ$] We utilize the attention weights to extract and visualize characteristic frames.
    \item[$\circ$] We carefully evaluate OperA on two challenging surgical video datasets.
\end{itemize} 

\section{Methodology}

\begin{figure}[t] 
	\centering
	\includegraphics[width=\textwidth]{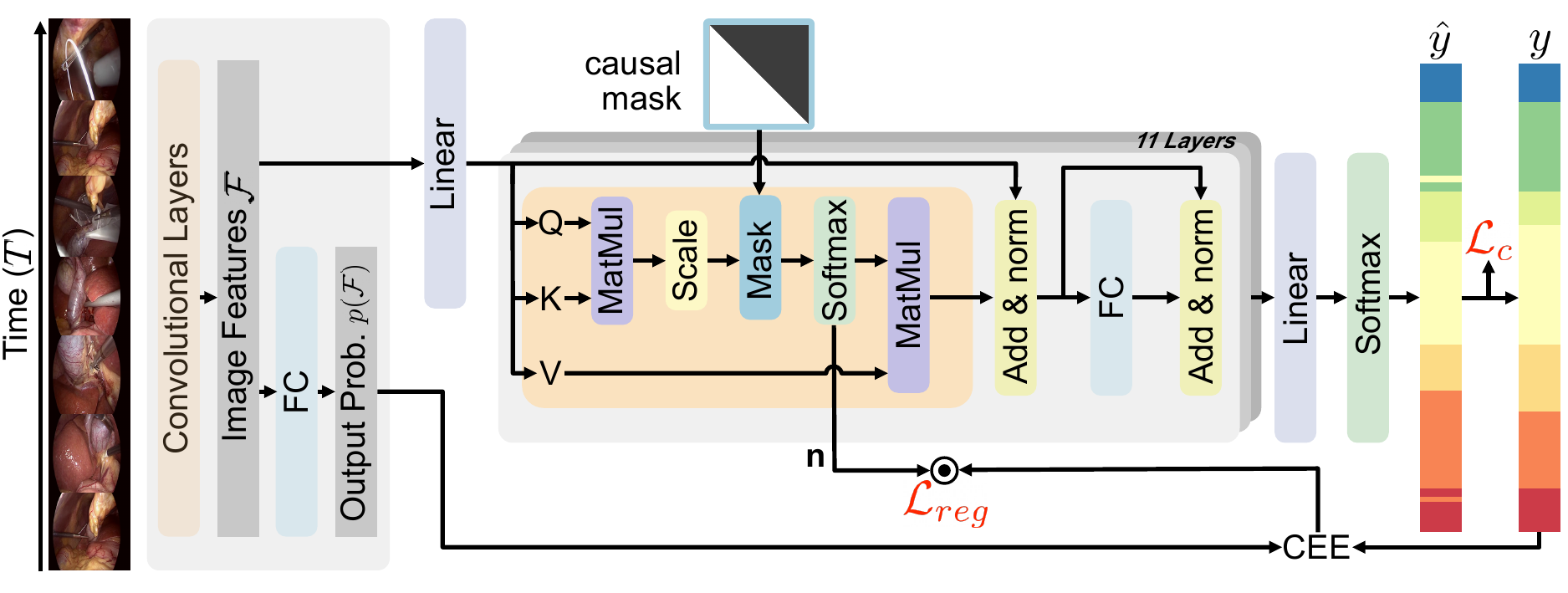}
	\caption{Overview of the proposed OperA model. Image features $\mathcal{F}$ are used as input for the transformer. The output logits $p(\mathcal{F})$ of the feature extraction backbone are used in combination with the normalized frame-wise attention weights \textbf{n} to regularize the attention.}
	\label{fig1}
\end{figure}

Our proposed model, OperA, consists of a CNN for visual feature extraction followed by multiple self-attention layers. The attention map is regularized during training to focus on reliable CNN image features. The full network architecture is visualized in Fig.~\ref{fig1}.

For our feature extraction backbone we trained a ResNet-50~\cite{He2016} frame-wise CNN without sequential modeling. We trained this model on phase recognition and additionally on surgical tool detection, if tool information was available in the dataset. The result of the feature extraction backbone are per frame image features $\mathcal{F} \in \mathbb{R}^{2048}$ and their corresponding class probabilities $p(\mathcal{F}) \in [0,1]^{c}$ with $c$ the number of classes.

 \subsection{Sequential Transformer Network} 
Our model expands on the well-known Transformer architecture~\cite{Vaswani2017} with the addition of our attention regularization that will be discussed below. Transformers  have the capabilities to model long sequences in a parallel manner using self-attention by relating every input feature with other input features regardless of their distance in the sequence~\cite{Kim2017}.
Visualized in Fig.~\ref{fig1}, we first calculate the query $Q$, key $K$ and value $V$, the inputs for the scaled dot product attention using a linear layer such that $\left( Q,K,V \right) = \text{Linear} \left(\mathcal{F} \right) \in \mathbb{R}^{3d}$ with $d=64$.

 \begin{equation}
    \text{AttentionWeights}(Q,K) =   \text{softmax} \left(\text{mask}\left( \frac{QK^{T}}{\sqrt{d}} \right) \right)  
    \label{eq_attentionweights}
 \end{equation}
 
 \begin{equation}
          \text{Attention}(Q,K,V) = \text{AttentionWeights}(Q,K)V
 \end{equation}

Our architecture uses 11 consecutive layers each consisting of a linear layer, a scaled dot-product attention layer, a layer normalization~\cite{Ba2016} and residual connections~\cite{He2016}.
Similar to the architecture of the Vision Tranformer~\cite{Dosovitskiy2020} after the last encoding layer, a linear layer followed by a softmax is used to estimate the class-wise output probabilities $y$ for each frame of the sequence.
For the training of the model we use a median frequency balanced cross-entropy loss~\cite{Eigen2015} $\mathcal{L}_{c}$. Causal masking~\cite{Rfou2019} with the binary mask $M \ni \{0, 1\}$ is performed on the attention map of the model, to prevent information leakage from future frames to the current frame prediction. This allows us to use OperA for real-time surgical phase prediction.

\subsection{Normalized Frame-Wise Attention}\label{Normalized frame wise attention weight}
For each frame, each layer of our transformer generates one column in $A = \text{AttentionWeights}(Q,K).$
Due to the softmax activation function (Eq.~\ref{eq_attentionweights}) each row of the attention map $A$ sums up to 1. Column-wise summation of the attention weights in $A$ results in the total attention value for each frame at time $t$. Due to the causal mask visualized in Fig.~\ref{fig1}, the first frame has the opportunity to contribute $T$ times, where $T$ is the number of frames in a video, while the last frame is considered only once.
We therefore need to normalize the frame-wise attention by dividing the total attention of each frame with the number of times this frame is considered, thus equally weighting the attention of all frames regardless of their position in the video.
The normalized frame-wise attention is calculated by: $\textbf{n} = \left( n_1, \ldots , n_T \right)$ with $n_{j} = \dfrac{\sum_i A_{ij}}{\sum_i M_{ij}}$.

\subsection{Attention Regularization}\label{attnreg}
Different from NLP and Visual Transformers, the input of OperA is generated by a CNN backbone network. The quality of each frame embedding of the CNN can vary drastically, especially for frames where the CNN predictions are incorrect. To this end, OperA should focus on higher quality CNN features, that were correctly classified by the backbone CNN. Such features have higher softmax probabilities, or confidence, and lower cross-entropy values.

We learn this relationship by comparing the normalized frame-wise attention weights with the prediction error of our CNN.
The regularization then reads:
 \begin{equation}
     \mathcal{L}_{reg} = \langle\textbf{n}, \text{CEE} \left( p(\mathcal{F}), y \right) \rangle
 \end{equation}

The Cross Entropy Evaluation value (CEE) describes the residual error of $p(\mathcal{F})$ compared to the ground truth label $y$. It should be noted that the weights of the backbone CNN remain frozen and CEE is only used for the optimization of the attention weights.
Multiplying CEE with the normalized frame-wise attention $\textbf{n}$ explicitly penalizes the model if a high attention value was generated for a feature with low CNN confidence. We apply the proposed regularization to the first attention layer as it has a direct relationship with the input visual features.
The final loss function used for model training is denoted as: $\mathcal{L} = \mathcal{L}_{c} + \lambda \cdot \mathcal{L}_{reg}
     \label{overallloss}$
     
Summing up all the normalized attention weights for each layer we generate a final attention value for each frame. In order to interpret whether OperA focuses on highly-informative frames that correctly represent each phase, we extract the frames with Highest Attention (HA) and the ones with Lowest Attention (LA).
\section{Experimental Setup}
\subsubsection{Datasets}
For the evaluation of OperA we use two challenging surgical workflow intra-operative video datasets of laparoscopic cholecystectomy procedures. The publicly available Cholec80~\cite{Twinanda2017} includes 80 videos with a resolutions of 1920$\times$1080 or 854$\times$480 pixels recorded at 25 frames-per-second (fps). For this work, the dataset was sub-sampled to 1fps. Every frame in the video has been manually assigned to one out of seven surgical phases. Additionally, seven different tool annotation labels sampled at 1fps are provided. We randomly select 20 videos for testing and the remaining 60 videos for training (48) and validation (12). 

MitiSW (Miti Surgical Workflow) was collected and annotated by the MITI group at the Klinikum rechts der Isar in Munich. The dataset consists of 85 laparoscopic cholecystectomy videos with resolution 1920$\times$1080 pixels and sampling rate of 1fps. MitiSW includes the 7 surgical phases of Cholec80, shown in Fig.~\ref{fig3}, along with one additional phase Pre-preparation, used to describe frames before the Preparation phase.
The phases have been annotated by expert physicians with no additional tool-presence information. 20 videos are utilized for testing and the remaining 65 videos for training (52) and validation (13).
For all the experiments 5-fold cross validation is performed. To balance our combined loss function we set $\lambda$ to $1$.

\subsubsection{Model Training}
OperA was trained for the task of surgical phase recognition using the Adam optimizer with an initial learning rate of 1e-5 for 30 epochs.
We report the test results extracted by the model that performed best on the validation set for each fold. The batch size is identical to the length of each video. Our method was implemented in PyTorch and our models were trained on an NVIDIA Titan V 12GB GPU using Polyaxon\footnote{https://polyaxon.com/}. The source code for OperA along with the Evaluation scripts is publicly available\footnote{https://github.com/tobiascz/OperA/}.
\begin{table}[t]
    \centering
    \caption{Ablative testing results for 6 and 11 transformer layers and with the addition of Attention Regularization (Reg). Average metrics over 5 folds are reported (\%) with the corresponding standard deviation ($\pm$).}
    \begin{tabu} to \textwidth { X[c] X[c] X[2c] X[2c] X[2c] X[2c] }
        \multicolumn{2}{c}{} & \multicolumn{2}{c}{\textbf{Cholec80}} & \multicolumn{2}{c}{\textbf{MitiSW}} \\ 
        \cmidrule(){3-6}
         {Layers} & {Reg} & \centering{Acc} & \centering{F1} & \centering{Acc} & \centering{F1}  \\
        \cmidrule(){1-2} \cmidrule(lr){3-4} \cmidrule(l){5-6} 
         {6} & {-} & 90.35 $\pm$ 0.71   & 83.45 $\pm$ 0.32 & 84.88 $\pm$ 1.43 & 84.92 $\pm$ 1.12 \\
         {6} & {\checkmark} & 90.49 $\pm$ 0.70   & 84.01 $\pm$ 0.39 & 85.41 $\pm$ 1.26 & 85.14 $\pm$ 1.20 \\
         {11} & {-} & 90.37 $\pm$ 0.86   & 83.85 $\pm$ 0.33 & 85.02 $\pm$ 1.01 & 85.28 $\pm$ 0.98 \\
         {11} & {\checkmark} & \textbf{91.26} $\pm$ \textbf{0.64}   & \textbf{84.49} $\pm$ \textbf{0.60} & \textbf{85.77} $\pm$ \textbf{0.95} & \textbf{85.44} $\pm$ \textbf{0.79} \\
    \end{tabu}
    \label{tab:ablative_testing}
\end{table}
\subsubsection{Evaluation Metrics and Baselines}
To comprehensively measure the results we report the video-level Accuracy (Acc) the harmonic mean (F1) of Precision and Recall~\cite{Padoy2012} and average the results over the 5 splits. We perform ablative testing to identify the most suitable number of attention layers and to test the effect of the regularization term (Sec.~\ref{attnreg}). Finally, we compare OperA with a variety of surgical phase recognition baselines.  
\section{Results and Discussion} \label{results}

\subsubsection{Effect of Layers and Regularization}
In Table~\ref{tab:ablative_testing} we compare models trained with 6 and 11 attention layers and evaluate the impact of the attention regularization. 11 attention layers is the highest amount of layers that we could fit in the VRAM. For both datasets, the results slightly increase for 11 attention layers by $\sim$1\% both in terms of Accuracy and F1-score. Regarding the attention regularization, a $\sim$1\% improvement is reported for both datasets and number of layers. As we will discuss later, the attention regularization does not only marginally increases the model performance but also the quality of the highest attention video frames. 

\subsubsection{Baseline Comparison}
We compare OperA with various methods for surgical phase recognition. ResNet-50 is the feature extraction backbone, ResLSTM~\cite{Jin2018} and MTRCNet-CL~\cite{Jin2020} utilize LSTMs for the temporal refinement. Different to the other models MTRCNet-CL is trained in an end-to-end fashion combining a CNN feature extraction with LSTM training. One of the downsides to this approach is that due to memory constraint only a limited sequence of the video can be used per-batch. ResLSTM, TeCNO and OperA use pre-trained image features and can therefore analyze a full video sequence at once.

\noindent First, we see that temporal refinement achieves a substantial improvement over ResNet-50 ranging from 4-10\% for Cholec80 and 9-12\% for MitiSW, showcasing its advantage. MTRCNet-CL is outperformed by the other temporal models by 2-6\% potentially potentially due to the limited sequence length that can be processed in every batch. OperA with or without positional encoding (PE)~\cite{Vaswani2017} outperforms the other temporal models by 2-6\% in terms of accuracy for Cholec80 showcasing the abilities of transformers to model long temporal dependencies. Regarding MitiSW, OperA without PE has increased accuracy by 0.6-3\% and F1-score by 1\% over all baselines. For both datasets it can be seen that PE marginally decreases the performance, potentially due to the increased sequence length in surgical videos in comparison to NLP tasks.
\begin{table}[t]
    \centering
    \caption{Baseline comparisons for Cholec80 and MitiSW. MTRCNet-CL  requires  tool  information,  therefore  cannot  be  used for  MitiSW.  We report the average metrics over 5-fold cross validation along with their respective standard deviation ($\pm$).}
    \begin{tabu} to \textwidth { X[1.1c] X[c] X[c] X[c] X[c] }
        %\toprule
         & \multicolumn{2}{c}{\textbf{Cholec80}} & \multicolumn{2}{c}{\textbf{MitiSW}} \\ 
        \cmidrule(){2-5}
          & \centering{Acc} & \centering{F1} & \centering{Acc} & \centering{F1}  \\
        \cmidrule(lr){2-3} \cmidrule(l){4-5} 
\textbf{ResNet-50} & 81.21 $\pm$ 1.16   & 72.98 $\pm$ 1.17 & 73.90 $\pm$ 1.89 & 71.53 $\pm$ 1.41 \\
\textbf{ResLSTM} & 87.94 $\pm$ 0.80   & 82.29 $\pm$ 0.78 & 82.97 $\pm$ 1.18 & 84.06 $\pm$ 1.15 \\
\textbf{MTRCNet-CL} & 85.64 $\pm$ 0.21 & 80.94 $\pm$ 0.95 & \textendash  &  \textendash  \\
\textbf{TeCNO} & 89.05 $\pm$ 0.79   & 84.04 $\pm$ 0.64 & 85.09 $\pm$ 1.67 & 84.18 $\pm$ 1.53 \\
\textbf{OperA + PE} & 90.20 $\pm$ 1.45 & 83.34 $\pm$ 0.97 & 83.67 $\pm$ 1.54 & 84.04 $\pm$ 1.20 \\
\textbf{OperA} & \textbf{91.26} $\pm$ \textbf{0.64} & \textbf{84.49} $\pm$ 0.64 & \textbf{85.77} $\pm$ \textbf{0.95} & \textbf{85.44} $\pm$ \textbf{0.79}
    \end{tabu}
    \label{tab:baseline_testing}
\end{table}

\begin{figure}[htbp] 
	\centering
	\includegraphics[width=\textwidth]{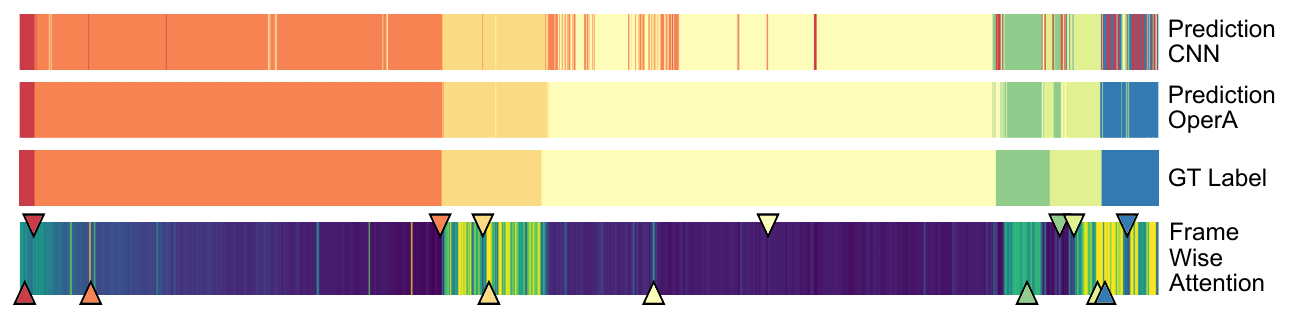}
	\caption{Qualitative results of the predictions per phase for video 66 from Cholec80 using the feature extraction CNN and OperA compared to the ground truth labels. In the frame-wise attention, brighter (yellow) color corresponds to higher attention, darker (blue) color to lower attention. The position of the LA frames for each phase is denoted with $\bigtriangledown$ and the HA frames with $\triangle$.}
	\label{fig2}
\end{figure}

\subsubsection{Predictions and Attention Values} In Fig.~\ref{fig2} we visualize the ground truth, predictions and attention values for video 66 of Cholec80. First, we see that the predictions of OperA are smoother and more consistent than the ones of the CNN. Moreover, the HA frames (denoted with $\triangle$) in most cases correspond with frames, where the CNN and OperA predictions were correct, while LA frames (denoted with $\bigtriangledown$) are at positions where the CNN predictions were wrong, confirming that attention regularization works as intended.

\begin{figure}[t] 
	\centering
	\includegraphics[width=\textwidth]{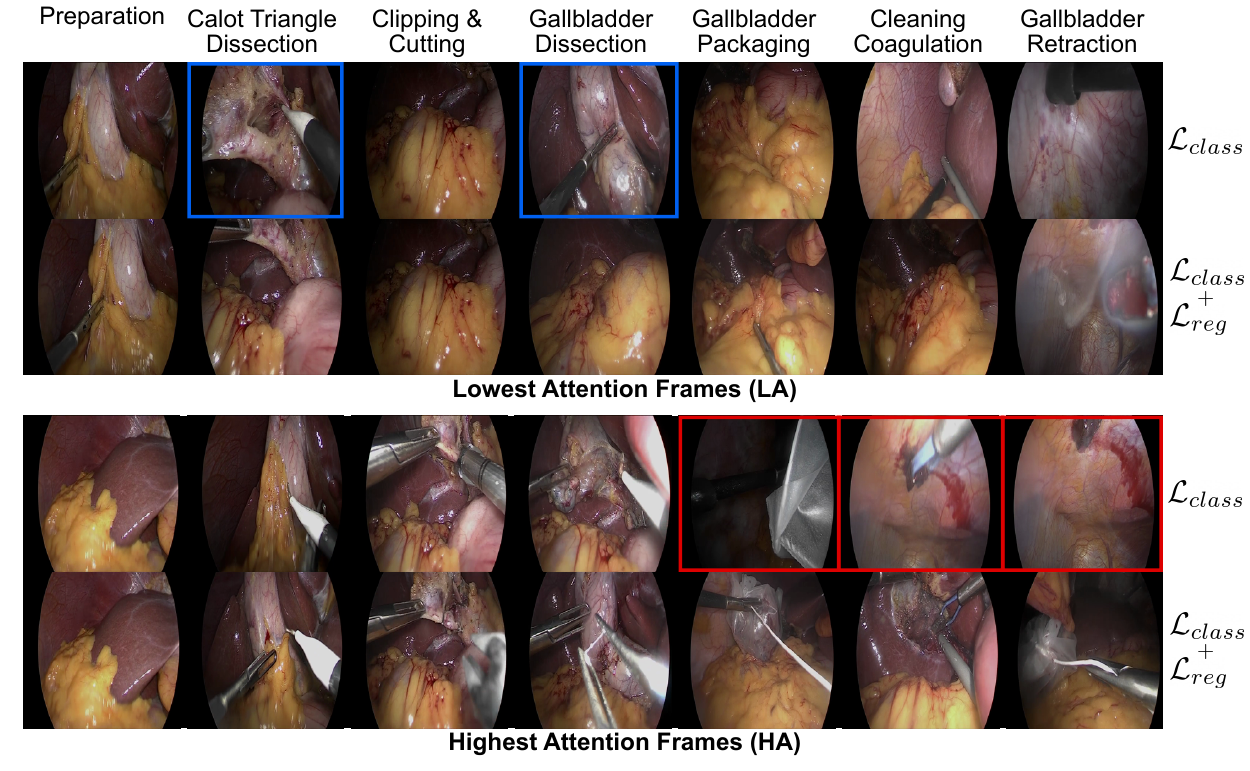}
	\caption{Visualization of frames of video 66 of Cholec80 with highest (HA) and lowest (LA) attention per phase for the models with and without attention regularization. Blue and red boxes denote frames of the model without regularization that have low attention, while they are descriptive of their phase and high attention, while they are not.}
	\label{fig3}
\end{figure}

\subsubsection{Highest and Lowest Attention Frames} In Fig.~\ref{fig3} we visualize the HA and LA frames per phase for the models trained with and without attention regularization. Visual inspection revealed that LA frames are generally less descriptive for the respective surgical phase. However, as we can see highlighted by the blue boxes, the model trained without attention regularization has minimum attention for frames containing surgical tools that are quite characteristic of their respective phase.

\noindent Regarding the HA frames, they are more diverse and representative of their surgical phase. For the model trained with attention regularization surgical tools are present in all phases besides ``Preparation'', highlighting the strong correlation between tools and surgical phases even though the attention model was not trained with tool information. 
With red boxes, we are showcasing the HA frames for the model trained without attention regularization. These frames were not descriptive of their phase and very similar to each other in the case of ``Cleaning Coagulation'' and ``Gallbladder Retraction''. These findings highlight the benefits of the proposed attention regularization and its potential for surgery video summarization.
\section{Conclusion}
In this paper we introduced OperA, a transformer-based model that accurately predicted surgical phases of cholecystectomy procedures, outperforming a variety of baselines on two challenging laparoscopic video datasets. Additionally, our novel attention regularizer enabled OperA to extract characteristic high attention frames. Future work includes applying our method to different laparoscopic procedures and explore its potential for surgical video summarization.

\section*{Acknowledgements}
\label{sec:acknowledgements}
\noindent
This work is partially funded by the DFG research unit PLAFOKON (FKZ: NA620/33-2) and BMBF project ARTEKMED (FKZ: 16SV8088) in collaboration with the Minimal-invasive Interdisciplinary Intervention Group (MITI). Finally, we would like to thank NVIDIA for the GPU donation.

\bibliographystyle{ieeetr}
\bibliography{bibfile}

\newpage
\section{Supplementary}

\begin{table}[h!]
    \begin{subtable}{1\textwidth}
    \centering
    \caption{Ablative testing results for 6 and 11 transformer layers and with the addition of Attention Regularization (Reg). Average metrics over 5 folds are reported (\%) with the corresponding standard deviation ($\pm$).}
    \begin{tabu} to \textwidth { X[c] X[c] X[2c] X[2c] X[2c] X[2c] }
        %\toprule
        {Layers} & {Reg} & \centering{Accuracy} & \centering{F1} & \centering{Precision} & \centering{Recall}  \\
        \cmidrule(r){1-2} \cmidrule(l){3-6}
        \multicolumn{2}{c}{} & \multicolumn{4}{c}{\textbf{Cholec80}} \\ 
         \cmidrule(lr){3-6}
         {6} & {-} & 90.35 $\pm$ 0.71   & 83.45 $\pm$ 0.32 & 80.64 $\pm$ 1.41 & 86.48 $\pm$ 0.61 \\
         {6} & {\checkmark} & 90.49 $\pm$ 0.70   & 84.01 $\pm$ 0.39 & 81.38 $\pm$ 0.29 & \textbf{86.98} $\pm$ \textbf{0.61} \\
         {11} & {-} & 90.37 $\pm$ 0.86   & 83.85 $\pm$ 0.33 & 81.60 $\pm$ 0.40 & 86.23 $\pm$ 0.34 \\
         {11} & {\checkmark} & \textbf{91.26} $\pm$ \textbf{0.64}   & \textbf{84.48} $\pm$ \textbf{0.60} & \textbf{82.19} $\pm$ \textbf{0.70} & 86.92 $\pm$ 0.86 \\ \\
         \multicolumn{2}{c}{} & \multicolumn{4}{c}{\textbf{MitiSW}} \\ 
         \cmidrule(){3-6}
         {6} & {-} & 84.88  $\pm$ {1.43} & 84.92  $\pm$ {1.12} & 82.76  $\pm$ {1.43} & 87.20  $\pm$ {1.02} \\
         {6} & {\checkmark} & 85.41 $\pm$ {0.95} & 85.14 $\pm$ {1.20} & 83.00 $\pm$ {1.34} & 87.41 $\pm$ {1.66} \\  
         {11} & {-} & 85.02  $\pm$ {1.01} & 85.28  $\pm$ {0.98} & 82.89  $\pm$ {1.20} & \textbf{87.82}  $\pm$ \textbf{0.75} \\
         {11} & {\checkmark} & \textbf{85.77}  $\pm$ \textbf{0.95} & \textbf{85.44}  $\pm$ \textbf{0.78} & \textbf{83.32}  $\pm$ \textbf{1.52} & 87.68  $\pm$ 1.08 \\

    \end{tabu}
    \label{tab:Full Ablative Testing}
    \end{subtable}

    \begin{subtable}{1\textwidth}
    
        \centering
    \caption{Baseline comparisons for Cholec80 and MitiSW. MTRCNET-CL  requires  tool  information,  therefore  cannot  be  used for  MitiSW.  We report the average metrics over 5-fold cross validation along with their respective standard deviation ($\pm$).}
    \begin{tabu} to \textwidth { X[2c] X[2c] X[2c] X[2c] X[2c] }
        %\toprule
        & \centering{Accuracy} & \centering{F1} & \centering{Precision} & \centering{Recall}  \\
        \cmidrule(){1-5}
        \multicolumn{1}{c}{} & \multicolumn{4}{c}{\textbf{Cholec80}} \\ 
        \cmidrule(lr){2-5}
        \textbf{ResNet-50} & 81.21 $\pm$ 1.16   & 72.98 $\pm$ 1.17 & 68.35 $\pm$ 1.61 & 78.31 $\pm$ 1.14 \\
        \textbf{ResLSTM} & 87.94 $\pm$ 0.80   & 82.29 $\pm$ 0.78 & 80.26 $\pm$ 1.12 & 84.43 $\pm$ 0.85 \\
        \textbf{MTRCNet-cl} & 85.64 $\pm$ 0.21 & 80.94 $\pm$ 0.95 & 79.31 $\pm$ 0.97 & 82.67 $\pm$ 0.114 \\
        \textbf{TeCNO} & 89.05 $\pm$ 0.79   & 84.04 $\pm$ 0.64 & 80.90 $\pm$ 0.75 & \textbf{87.44} $\pm$ \textbf{0.64} \\
        \textbf{OperA + PE} & 90.20 $\pm$ 1.45 & 83.34 $\pm$ 0.97 & 80.78 $\pm$ 1.42 & 86.08 $\pm$ 0.89 \\
        \textbf{OperA} & \textbf{91.26} $\pm$ \textbf{0.64}   & \textbf{84.48} $\pm$ \textbf{0.60} & \textbf{82.19} $\pm$ \textbf{0.70} & 86.92 $\pm$ 0.86 \\ \\
        \multicolumn{1}{c}{} & \multicolumn{4}{c}{\textbf{MitiSW}} \\ 
        \cmidrule(lr){2-5}
        \textbf{ResNet-50} & 73.90 $\pm$ 1.89 & 71.53 $\pm$ 1.41 & 69.06 $\pm$ 1.38 & 74.20 $\pm$ 1.63 \\
        \textbf{ResLSTM} & 82.97 $\pm$ 1.18   & 84.06 $\pm$ 1.15 & 82.08 $\pm$ 1.57 & 86.15 $\pm$ 0.94 \\
        \textbf{TeCNO} & 85.09 $\pm$ 1.67 & 84.18 $\pm$ 1.53 & 82.03 $\pm$ 0.20 & 86.50 $\pm$ 0.43 \\
        \textbf{OperA + PE} & 83.67 $\pm$ 1.54 & 84.04 $\pm$ 1.20 & 81.34 $\pm$ 1.60 & 86.94 $\pm$ 0.98 \\
        \textbf{OperA} & \textbf{85.77} $\pm$ \textbf{0.95}  & \textbf{85.44} $\pm$ \textbf{0.79} & \textbf{83.32} $\pm$ \textbf{1.10} & \textbf{87.68} $\pm$ \textbf{0.71 }
    \end{tabu}
    \label{tab:Full Baseline}
    \end{subtable}
\end{table}

\noindent
\begin{figure}[p] 
	\centering
	\includegraphics[width=\textwidth]{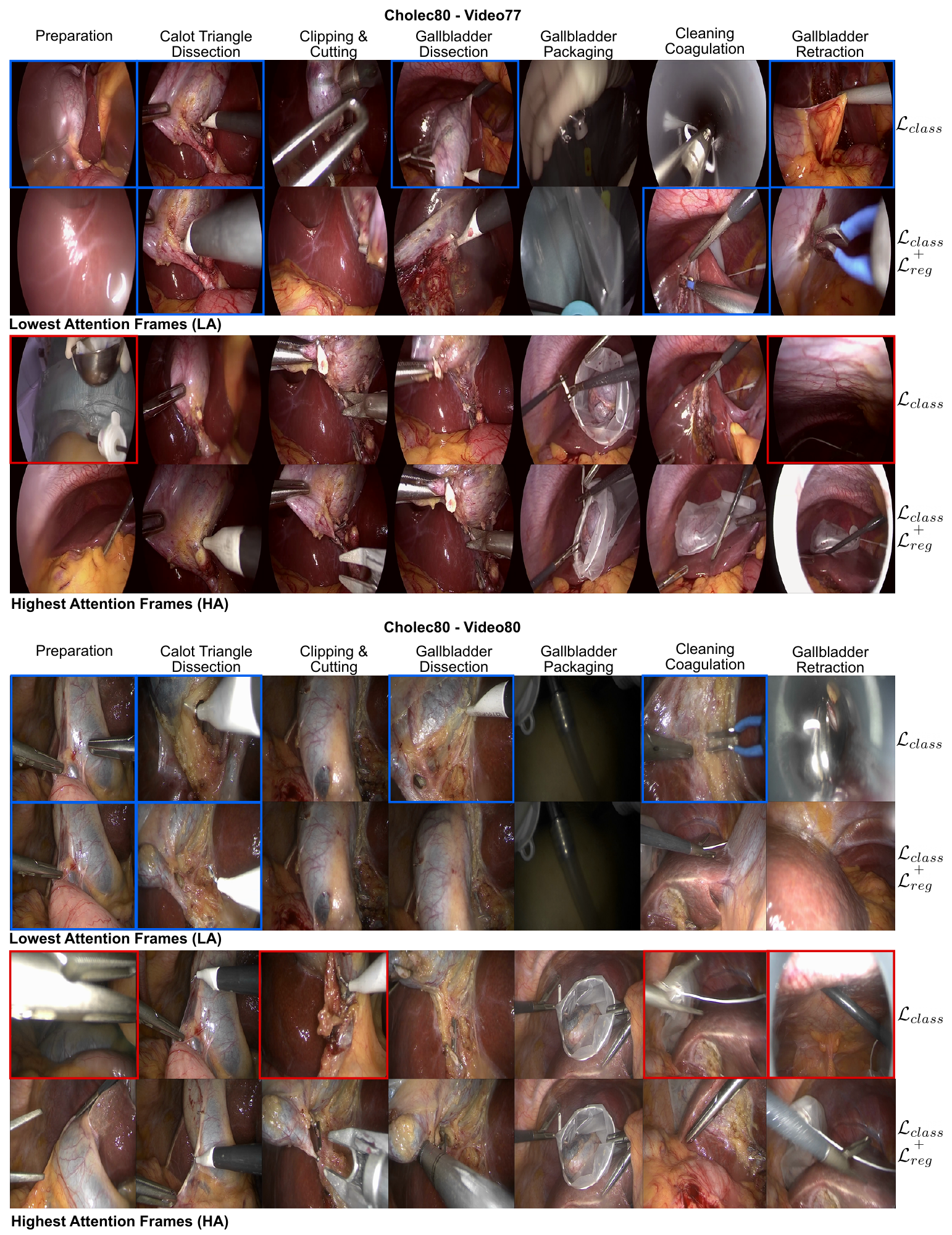}
	\caption{Two more videos from Cholec80 are visualized with maximum and minimum attention per phase for the models with and without attention regularization. Blue and red boxes denote frames of the model without regularization that have low attention, while they are descriptive of their phase and high attention, while they are not.}
	\label{supp_combined}
\end{figure}

\end{document}